\documentclass[
    a4paper,
    twoside,
]{article}

\usepackage{epsfig}
\usepackage{subcaption}
\usepackage{calc}
\usepackage{amssymb}
\usepackage{amstext}
\usepackage{amsmath}
\usepackage{amsthm}
\usepackage{multicol}
\usepackage{pslatex}
\usepackage{apalike}
\usepackage{algorithm2e}
\usepackage[bottom]{footmisc}

\usepackage{hyperref}
\usepackage[capitalize]{cleveref}
\usepackage{csquotes}
\usepackage{multirow}
\usepackage{siunitx}
\usepackage{booktabs}
\usepackage{nicefrac}
\usepackage{interval}
\usepackage[dvipsnames]{xcolor}

\usepackage{tikz}
\usetikzlibrary{arrows}
\usetikzlibrary{calc}
\usetikzlibrary{positioning}

\usepackage{pgfplots}
\pgfplotsset{compat=1.17}
\usepgfplotslibrary{external}
\pgfplotsset{%
    default/.style={%
        width=\columnwidth,
        grid=major,
        grid style={Gray!25},
        legend cell align={%
            left
        },
        legend style={%
            font=\footnotesize
        },
        ticklabel style={%
            font=\footnotesize,
        },
        mark size=1,
        mark options={solid, thick, scale=2},
        label style={rotate=0},
        enlargelimits=true,
        yticklabel style={%
            /pgf/number format/fixed,
            /pgf/number format/precision=2,
            /pgf/number format/fixed zerofill,
        },
        scaled y ticks=false,
    },
}

\hypersetup{%
    pdftoolbar=true,        % show Acrobat’s toolbar?
    pdfmenubar=true,        % show Acrobat’s menu?
    pdffitwindow=false,     % window fit to page when opened
    pdfstartview={FitH},    % fits the width of the page to the window
    pdftitle={Make Deep Networks Shallow Again},    % title
    pdfauthor={Bernhard Bermeitinger, Tomas Hrycej, Siegfried Handschuh},     % author
    pdfkeywords={Residual Connection} {Deep Neural Network} {Convolutional Network} {Computer Vision}, % list of keywords
    pdfnewwindow=true,      % links in new window
    breaklinks=true,
    colorlinks,
    linkcolor={red!50!black},
    citecolor={blue!50!black},
    urlcolor={blue!80!black}
}

\usepackage{SCITEPRESS}

\intervalconfig{%
    left closed fence=[,
    left open fence=(,    % chktex 36
    right closed fence=], % chktex 9
    right open fence=),   % chktex 9
}

\newcommand{\mnist}{\mbox{MNIST}}
\newcommand{\cifar}{\mbox{CIFAR10}}

\begin{document}

\title{Make Deep Networks Shallow Again}

\author{%
    \authorname{%
        Bernhard Bermeitinger\sup{1}\orcidAuthor{0000-0002-2524-1850},% chktex 8
        Tomas Hrycej\sup{1}%
        and Siegfried Handschuh\sup{1}\orcidAuthor{0000-0002-6195-9034}% chktex 8
    }
    \affiliation{%
        \sup{1}Institute of Computer Science, University of St.Gallen (HSG), St.Gallen, Switzerland
    }
    \email{bernhard.bermeitinger@unisg.ch, tomas.hrycej@unisg.ch, siegfried.handschuh@unisg.ch}
}


\abstract{%
    Deep neural networks have a good success record and are thus viewed as the best architecture choice for complex applications.
    Their main shortcoming has been, for a long time, the vanishing gradient which prevented the numerical optimization algorithms from acceptable convergence.
    A breakthrough has been achieved by the concept of residual connections---an identity mapping parallel to a conventional layer.
    This concept is applicable to stacks of layers of the same dimension and substantially alleviates the vanishing gradient problem.
    A stack of residual connection layers can be expressed as an expansion of terms similar to the Taylor expansion.
    This expansion suggests the possibility of truncating the higher-order terms and receiving an architecture consisting of a single broad layer composed of all initially stacked layers in parallel.
    In other words, a sequential deep architecture is substituted by a parallel shallow one.
    Prompted by this theory, we investigated the performance capabilities of the parallel architecture in comparison to the sequential one.
    The computer vision datasets \mnist{} and \cifar{} were used to train both architectures for a total of \num{6 912} combinations of varying numbers of convolutional layers, numbers of filters, kernel sizes, and other meta parameters.
    Our findings demonstrate a surprising equivalence between the deep (sequential) and shallow (parallel) architectures.
    Both layouts produced similar results in terms of training and validation set loss.
    This discovery implies that a wide, shallow architecture can potentially replace a deep network without sacrificing performance.
    Such substitution has the potential to simplify network architectures, improve optimization efficiency, and accelerate the training process.
}

\onecolumn \maketitle \normalsize \setcounter{footnote}{0} \vfill

% !TeX encoding = UTF-8
% !TeX spellcheck = en_US-large
% !TeX root = main.tex
\section{\uppercase{Introduction}}\label{sec:introduction}
Deep neural networks (i.e., networks with many nonlinear layers) are widely considered to be the most appropriate architecture for mapping complex dependencies such as those arising in Artificial Intelligence tasks.
Their potential to map intricate dependencies has advanced their widespread use.

For example, the study~\cite{meir2023EfficientShallowLearning} compares the first deep convolutional network for image classification with two sequential convolutional layers \emph{LeNet}~\cite{lecun1989BackpropagationAppliedHandwritten} to its deeper evolution \emph{VGG16}~\cite{simonyan2015VeryDeepConvolutional} with 13 sequential convolutional layers.
While the performance gain in this comparison was significant, further increasing the depth resulted in very small performance gains.
Adding three additional convolutional layers to \emph{VGG16} improved the validation error slightly from \SI{25.6}{\percent} to \SI{25.5}{\percent} on the \mbox{ILSVRC-2014} competition dataset~\cite{russakovsky2015ImageNetLargeScale}, while increasing the number of trainable parameters from \num{138}M to \num{144}M.

However, training these networks remains a significant challenge, often navigated through numerical optimization methods based on the gradient of the loss function.
In deeper networks, the gradient can significantly diminish particularly for parameters distant from the output, leading to the well-documented issue known as the \enquote{vanishing gradient}.

A breakthrough in this challenge is the concept of \emph{residual connections}: using an identity function parallel to a layer~\cite{he2016DeepResidualLearning}.
Each residual layer consists of an identity mapping copying the layer's input to its output and a conventional weighted layer with a nonlinear activation function.
This weighted layer represents the residue after applying the identity.
The output of the identity and the weighted layer are summed together, forming the output of the residual layer.
The identity function plays the role of a bridge---or \enquote{highway}~\cite{srivastava2015TrainingVeryDeep-NIPS}---transferring the gradient w.r.t.\ layer output into that of the input with unmodified size.
In this way, it increases the gradient of layers remote from the output.

The possibility of effectively training deep networks led to the widespread use of such residual connection networks and to the belief that this is the most appropriate architecture type~\cite{mhaskar2017WhenWhyAre}.
However, extremely deep networks such as \emph{\mbox{ResNet-1000}} with ten times more layers than \emph{\mbox{ResNet-101}}~\cite{he2016DeepResidualLearning} often demonstrate a performance decline.

Although there have been suggestions for wide architectures like \emph{\mbox{EfficientNet}}~\cite{tan2019EfficientNetRethinkingModel}, these are still considered \enquote{deep} within the scope of this paper.

This paper questions the assumption that deep networks are inherently superior, particularly considering the persistent gradient problems.
Success with methods like residual connections can be mistakenly perceived as validation of the superiority of deep networks, possibly hindering exploration into potentially equivalent or even better-performing  \enquote{shallow} architectures.

To avoid such premature conclusions, we examine in this paper the relative performance of deep networks over shallow ones, focusing on a parallel or \enquote{shallow} architecture instead of a sequential or \enquote{deep} one.
The basis of the investigation is the mathematical decomposition of the mapping materialized by a stack of convolutional residual networks into a structure that suggests the possibility of being approximated by a shallow architecture.
By exploring this possibility, we aim to stimulate further research, opening new avenues for AI architecture exploration and performance improvement.

\section{\uppercase{Decomposition of stacked residual connections}}\label{sec:decomposition}
A layer of a conventional multilayer perceptron can be thought of as a mapping $ y = F_h \left( x \right) $.
With the residual connection concept~\cite{he2016DeepResidualLearning}, this mapping is modified to
\begin{equation}\label{eq:res_con0}
    y = Ix + F_h \left( x \right)
\end{equation}
For the $h$-th hidden layer, the recursive relationship is
\begin{equation}\label{eq:res_conH}
    z_h = I z_{h-1} + F_h \left( z_{h-1} \right)
\end{equation}
For example, the second and the third layers can be expanded to
\begin{equation}\label{eq:res_con2}
    z_2 = I z_1 + F_2 \left( z_1 \right)
\end{equation}
and
\begin{equation}\label{eq:res_con3}
\begin{aligned}
    z_3 &= I z_2 + F_3 \left( z_2 \right) \\
        &= I \left( I z_1 + F_2 \left( z_1 \right) \right) + F_3 \left(I z_1 + F_2 \left( z_1 \right) \right) \\
        &= I z_1 + F_2 \left( z_1 \right) + F_3 \left( I z_1 + F_2 \left( z_1 \right) \right)
\end{aligned}
\end{equation}
In the operator notation, it is
\begin{equation}\label{eq:res_conOp0}
    z_h = z_{h-1} + F_h * z_{h-1}
        = \left( I + F_h \right) * z_{h-1}
\end{equation}
For linear operators, the recursion up to the final output vector $y$ can be explicitly expanded~\cite[Section~6.7.3.1]{hrycej2023MathematicalFoundationsData}
\begin{equation}\label{eq:res_conOpH}
    y = \quad I * x
        + \sum^{H}_{h=1} F_h * x
        + \sum^{H}_{h=1} \sum^{H}_{k=1,k>h} F_k * F_h * x
    \cdots
\end{equation}
with all combinations of operator triples, quadruples, etc.\ up to the product of all $ H $ layer operators.

Typically, these layer mappings are not linear due to their activation functions such as \emph{sigmoid}, \emph{tanh}, or \emph{ReLU}.
As a result, it does not satisfy the condition $ F_h \left( x + z \right) = F_h \left( x \right ) + F_h \left( z \right) $.
However, their gradient is a linear operator.
In a multilayer perceptron with a residual connection, the error gradient w.r.t.\ the output of the $h$-th layer is
\begin{equation}\label{eq:res_conGradZ}
    \begin{aligned}
        \frac{\partial E}{\partial z_h}
        &= \left( \prod_{k=h+1}^{H} \frac{\partial z_k}{\partial z_{k-1}} \right) \frac{\partial E}{\partial z_H} \\
        &= \left( \prod_{k=h+1}^{H} \left( I + W_k^T \nabla F_k \right) \right) \frac{\partial E}{\partial z_H}
    \end{aligned}
\end{equation}
The error gradient w.r.t.\ the weights is, for both standard layers and those with residual connection
\begin{equation}\label{eq:res_conGradW}
    \frac{\partial E}{\partial W_h}
    = \nabla F_h \frac{\partial E}{\partial z_h} z^T_{h-1}
\end{equation}
and w.r.t.\ biases
\begin{equation}\label{eq:res_conGradB}
    \frac{\partial E}{\partial b_h}
    = \nabla F_h \frac{\partial E}{\partial z_h}
\end{equation}

This shows that the expansion given in~\cref{eq:res_conOpH} is valid for an approximation linearized with the help of the local gradient.
In particular, it is valid around the minimum.

In an analogy to Taylor expansion, it can be hypothesized that the first two terms
\begin{equation}\label{eq:res_conOpH2}
    y = I * x + \sum^{H}_{h=1} F_h * x
\end{equation}
may be a reasonable approximation of the whole mapping in~\cref{eq:res_conOpH}.

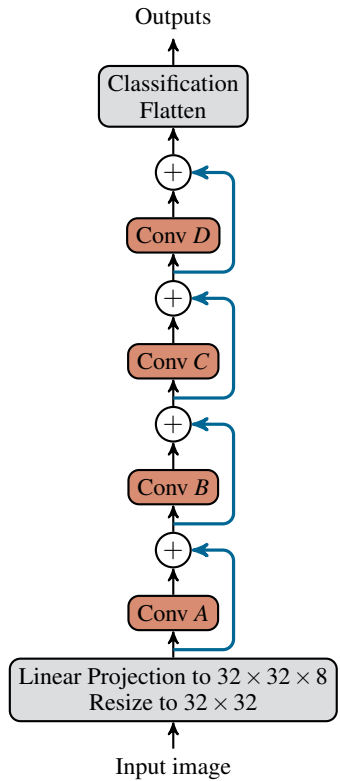
\begin{figure}
    \centering
    \tikzsetnextfilename{Network-Sequential}
\begin{tikzpicture}[
    font=\footnotesize,
    layer/.style={draw, thick, rounded corners, minimum width=1em},
    input/.style={layer},
    preprocessor/.style={layer, fill=Black!15},
    conv/.style={layer, fill=BrickRed!50},
    clf/.style={layer, fill=Black!15},
    connection/.style={-stealth', thick, rounded corners},
    skip/.style={connection, very thick, MidnightBlue},
    plus/.style={draw, thick, circle, font=\normalsize, inner sep=1pt},
    node distance=3.5mm,
]
% INPUT LAYERS
\node (inputs) {Input image};
\node[above=of inputs, preprocessor, align=center] (preprocessor) {Linear Projection to $ 32 \times 32 \times 8 $\\ Resize to $ 32 \times 32 $};

% CONVOLUTIONAL LAYERS
\node[above=of preprocessor, conv, align=center] (conv1) {Conv $A$};
\node[above=of conv1, plus] (plusConv1) {$+$};
\node[above=of plusConv1, conv, align=center] (conv2) {Conv $B$};
\node[above=of conv2, plus] (plusConv2) {$+$};
\node[above=of plusConv2, conv, align=center] (conv3) {Conv $C$};
\node[above=of conv3, plus] (plusConv3) {$+$};
\node[above=of plusConv3, conv, align=center] (conv4) {Conv $D$};
\node[above=of conv4, plus] (plusConv4) {$+$};

% CLASSIFICATION AND OUTPUT LAYERS
\node[above=of plusConv4, clf, align=center] (clf) {Classification\\ Flatten};
\node[above=of clf] (outputs) {Outputs};

% DATA FLOW FOR LAYERS
\draw[connection] (inputs) -- (preprocessor);
\draw[connection] (preprocessor) -- (conv1);
\draw[connection] (conv1) -- (plusConv1);
\draw[connection] (plusConv1) -- (conv2);
\draw[connection] (conv2) -- (plusConv2);
\draw[connection] (plusConv2) -- (conv3);
\draw[connection] (conv3) -- (plusConv3);
\draw[connection] (plusConv3) -- (conv4);
\draw[connection] (conv4) -- (plusConv4);
\draw[connection] (plusConv4) -- (clf);
\draw[connection] (clf) -- (outputs);

% DATA FLOW FOR SKIP CONNECTIONS
\coordinate (skipConv1) at ($(preprocessor.north)!.3!(conv1.south)$);
\coordinate [right=of plusConv1, xshift=.2cm] (plusConv1Right);
\coordinate (skipConv2) at ($(plusConv1.north)!.3!(conv2.south)$);
\coordinate [right=of plusConv2, xshift=.2cm] (plusConv2Right);
\coordinate (skipConv3) at ($(plusConv2.north)!.3!(conv3.south)$);
\coordinate [right=of plusConv3, xshift=.2cm] (plusConv3Right);
\coordinate (skipConv4) at ($(plusConv3.north)!.3!(conv4.south)$);
\coordinate [right=of plusConv4, xshift=.2cm] (plusConv4Right);

% SKIP CONNECTIONS
\draw[skip] (skipConv1) -| (plusConv1Right) -- (plusConv1);
\draw[skip] (skipConv2) -| (plusConv2Right) -- (plusConv2);
\draw[skip] (skipConv3) -| (plusConv3Right) -- (plusConv3);
\draw[skip] (skipConv4) -| (plusConv4Right) -- (plusConv4);

\end{tikzpicture}
    \caption{Overview of the sequential architecture with four consecutive convolutional layers with eight filters each and their skip connections.}\label{fig:example-network:sequential}
\end{figure}

\begin{figure}
    \centering
    \tikzsetnextfilename{Network-Parallel}
\begin{tikzpicture}[
    font=\footnotesize,
    layer/.style={draw, thick, rounded corners, minimum width=1em},
    input/.style={layer},
    preprocessor/.style={layer, fill=Black!15},
    conv/.style={layer, fill=BrickRed!50},
    clf/.style={layer, fill=Black!15},
    connection/.style={-stealth', thick, rounded corners},
    skip/.style={connection, very thick, MidnightBlue},
    plus/.style={draw, thick, circle, font=\normalsize, inner sep=1pt},
    node distance=4mm,
]
% INPUT LAYERS
\node (inputs) {Input image};
\node[above=of inputs, preprocessor, align=center] (preprocessor) {Linear Projection to $ 32 \times 32 \times 8 $\\ Resize to $ 32 \times 32 $};

% CONVOLUTIONAL LAYERS
\coordinate[above=8mm of preprocessor] (middle);
\node[right=2mm of middle, conv] (conv3) {Conv $C$};
\node[right=4mm of conv3, conv] (conv4) {Conv $D$};
\node[left=2mm of middle, conv] (conv2) {Conv $B$};
\node[left=4mm of conv2, conv] (conv1) {Conv $A$};
\coordinate[right=of conv4] (rightOfConv4);

% CLASSIFICATION AND OUTPUT LAYERS
\node[above=of conv3, plus] (plusConvs) {$+$};
\node[above=of plusConvs, clf, align=center] (clf) {Classification\\ Flatten};
\node[above=of clf] (outputs) {Outputs};

% DATA FLOW FOR LAYERS
\draw[connection] (inputs) -- (preprocessor);
\draw[connection] (preprocessor.north) -- ++(0,2mm) -| (conv1.south);
\draw[connection] (preprocessor.north) -- ++(0,2mm) -| (conv2.south);
\draw[connection] (preprocessor.north) -- ++(0,2mm) -| (conv3.south);
\draw[connection] (preprocessor.north) -- ++(0,2mm) -| (conv4.south);
\draw[connection] (conv1.north) -- (plusConvs.west);
\draw[connection] (conv2.north) -- (plusConvs.south west);
\draw[connection] (conv3.north) -- (plusConvs.south);
\draw[connection] (conv4.north) -- (plusConvs.south east);
\draw[connection] (plusConvs) -- (clf);
\draw[connection] (clf) -- (outputs);

% SKIP CONNECTIONS
\coordinate [right=of plusConvs, xshift=.2cm] (plusConvsRight);
\draw[skip]
    (preprocessor.north)
    -- ++(0,2mm)
    -| (rightOfConv4.south)
    -- (rightOfConv4.north)
    -- ++(0,8mm)
    -- (plusConvs.east)
    ;
\end{tikzpicture}
    \caption{Overview of the parallelized architecture of~\cref{fig:example-network:sequential} with four convolutional layers with eight filters each and one skip connection.}\label{fig:example-network:parallel}
\end{figure}
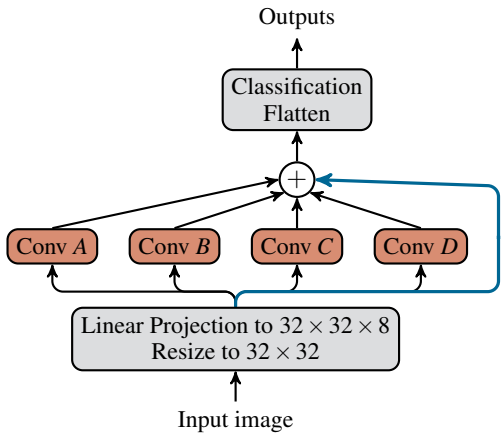

In terms of implementation as neural networks, the stack of layers with residual connections (as exemplified in~\cref{fig:example-network:sequential}) could be approximated by the parallel architecture such as that illustrated in~\cref{fig:example-network:parallel}.

Of course, this hypothesis has to be confirmed by tests on real-world problems.
If acceptable, it would be possible to substitute a deep residual network of $ H $ sequential layers with a \enquote{shallow} network with a single layer consisting of $ H $ individual modules in parallel, summing their output vectors.
Each of these modules would be equivalent to one layer in the deep architecture.
The main objective is not to prove that both networks are nearly equivalent with the same parameter set, as this is unlikely to be the case.
Rather, the goal is to demonstrate that both shallow and deep architectures can effectively learn and attain comparable performances on the given task.
The consequence would be that the shallow architecture can reach the same performance as the deep one, with the same number of parameters.
This may be relevant for the preferences in setting up neural networks for particular tasks since shallow networks suffer less from numerical computing problems such as vanishing gradient.

\section{\uppercase{Setup of computing experiments}}\label{sec:setup}
The analysis of~\cref{sec:decomposition} suggests that the expressive power of a network architecture in which stacked residual connection layers of a deep network are reorganized into a parallel operation in a single, broad layer, may be close to that of the original deep network.
This hypothesis is to be tested on practically relevant examples.

It is important to point out that residual connection layers are restricted to partial stacks of equally sized layers (otherwise the unity mapping could not be implemented).
A typical use of such networks is image classification where an image is processed by consecutive layers of size equal to the (possibly reduced) pixel matrix.
The output of this network is usually a vector of class probabilities that differ in dimensionality from that of the input image.
This is the reason for one or more non-residual layers at the output and some preprocessing non-residual layers at the input.

Residual connections can be used for any stack of layers of the same dimensions. However, in domains such as image processing, the layers are mostly of the \emph{convolutional} type.
This is a layer concept in which the same, relatively small weight matrix, is applied to the neighbor environment of every position in the input. They are implementing a local operator (such as edge detection) shifted over the extension of the image.
The following benchmark applications are using convolutional layers.

\emph{Filters} are a concept in convolutional layers which consist of a multiplicity of such convolution operators.
Each filter convolves individually with the input matrix for generating the output.
Multiple filters in a layer operate independently from each other, building a parallel structure.
The computing experiments reported here were done both with and without multiple filters.
The possibility of making the consecutive layer stack parallel concerns only the middle part with residual connections of identically sized layers.

For the experiments, the two well-known image classification datasets \mnist~\cite{lecun1998GradientbasedLearningApplied} and \cifar~\cite{krizhevsky2009LearningMultipleLayers} were used.
\mnist{} contains black and white images of handwritten digits (0--9) while \cifar{} contains color images of exclusively ten different mundane objects like \enquote{horse}, \enquote{ship}, or \enquote{dog}.
They contain \num{60 000} (\mnist{}) and \num{50 000} (\cifar{}) training examples.
Their respective preconfigured test split of each \num{10 000} examples are used as validation sets.
While \cifar{} is evenly distributed among all classes, \mnist{} is roughly evenly distributed with a standard deviation of \num{322} for the training set and \num{59} for the validation set.
We took no special treatment for this small class imbalance.

A series of computing experiments of all the following possible architectures were run:
\begin{itemize}
    \item
        Number of convolutional layers: $1$, $2$, $4$, $8$, $16$, $32$
    \item
        Number of filters per convolutional layer: $1$, $2$, $4$, $8$, $16$, $32$
    \item
        Kernel size of a filter: $ 1 \times 1 $, $ 2 \times 2 $, $ 4 \times 4 $, $6 \times 6$, $ 8 \times 8 $, $ 16 \times 16 $
    \item
        Activation function of each convolutional layer: \emph{sigmoid}, \emph{ReLU}
\end{itemize}

\Cref{fig:example-network:sequential} shows the sequential architecture with depth $ 4 $ and $ 8 $ filters per convolutional layer.
For comparison, the parallelized version is shown in \cref{fig:example-network:parallel}.
The sizes of the filters' kernels are not shown because they don't interfere with the layout.

The images are resized to $32 \times 32$ pixels to match the varying kernel sizes.
For the summation of the skip connection and the convolutional layer to work out, they need to have the same dimensionality.
Therefore, for preprocessing, the images are linearly mapped to match the convolutional layers' output dimensions.
To keep the architecture simple and reduce the possibility of additional side effects, the input is flattened into a one-dimensional vector before the dense classification layer with ten linear output units.
These linear layers are initialized with the same set of fixed random values throughout all experiments.

The same configuration setup was used for the number of parallel filters per layer.
Parallel filters are popular means of extending a straightforward convolution layer architecture: instead of each layer being a single convolution of the previous layer, it consists of multiple convolution filters in parallel.
In all well-performing image classifiers based on convolutional layers, multiple filters are used~\cite{fukushima1980NeocognitronSelforganizingNeural,krizhevsky2012ImageNetClassificationDeep,simonyan2015VeryDeepConvolutional}.

Throughout all experiments, the parameters of the layers at the same depths were always initialized with the same random values with a fixed seed.
For example, the two layers labeled $ A $ in~\cref{fig:example-network:sequential,fig:example-network:parallel} started their training from the same parameter set.

The categorical cross-entropy loss was employed as the loss function due to its suitability for multi-label classification problems.
This loss served also as the main assessment of the training performance.
An alternative would have been the most popular (and the most meaningful from the application point of view) metric: classification accuracy.
However, it would be a methodological fault to use a metric that is different from the loss function that is genuinely optimized.
The relationship between cross-entropy loss and classification accuracy is loaded with random effects and is frequently not even monotonic.
This justifies the selection of cross-entropy loss for performance review.

The batch size was set to \num{512}.
The datasets were not shuffled between epochs or experiments, leading to identical batches throughout all experiment runs.

As the optimizer, RMSprop~\cite{hinton2012NeuralNetworksMachine} was chosen with a fixed learning rate.
All experiments were duplicated for the learning rates \num{1e-2}, \num{1e-3}, \num{1e-4}, and \num{1e-5}.
Different learning rates had only a marginal effect on the results.
The figures and tables show the results obtained with a learning rate of \num{1e-4}.

Each experiment ran for \num{100} epochs, which resulted in \num{11 800}~optimization steps for \mnist{}, and \num{9 800}~steps for \cifar{}.
The \num{6 912} experiments were run individually on \emph{NVIDIA Tesla V100} GPUs for a total run time of \num{79}~days.
The results are reported for kernel size $ 16~\times~16 $ which showed the best average classification performance although not significantly different.

\section{\uppercase{Computing experiments}}\label{sec:experiments}

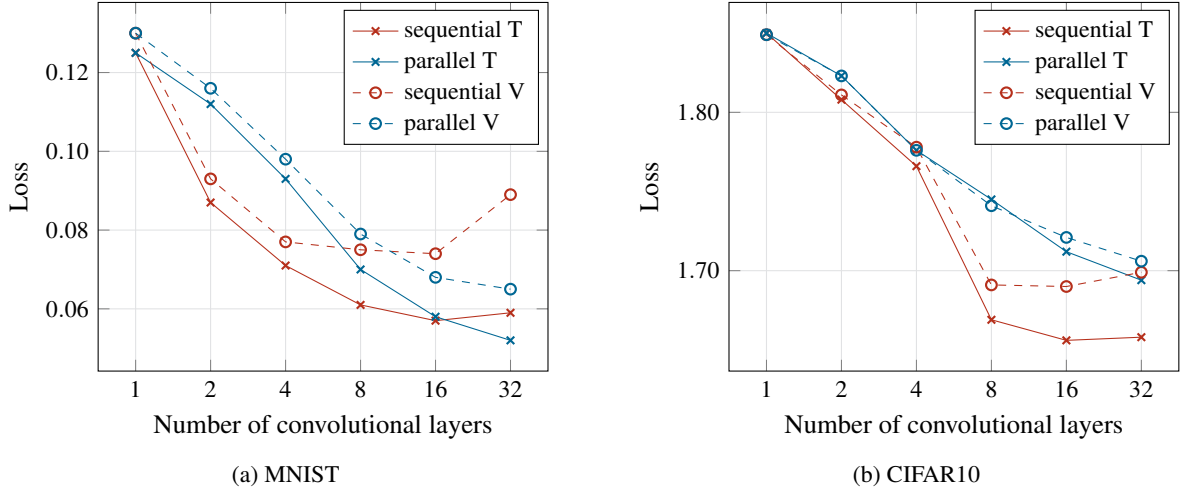
\begin{figure*}[ht]
    \centering
    \begin{subfigure}{\columnwidth}
        \tikzsetnextfilename{F1-mnist}
\begin{tikzpicture}
    \begin{semilogxaxis}[
        default,
        xlabel={Number of convolutional layers},
        ylabel={Loss},
        xtick={1,2,4,8,16,32},
        xticklabels={1,2,4,8,16,32},
    ]
        \addplot[
            solid, color=BrickRed, mark=x,
        ] table [
            x=x-mnist-F1-sequential-train, y=y-mnist-F1-sequential-train
        ] {data/F1.data};
        \addlegendentry{sequential T}

        \addplot[
            solid, color=MidnightBlue, mark=x,
        ] table [
            x=x-mnist-F1-parallel-train, y=y-mnist-F1-parallel-train
        ] {data/F1.data};
        \addlegendentry{parallel T}

        \addplot[
            dashed, color=BrickRed, mark=o,
        ] table [
            x=x-mnist-F1-sequential-val, y=y-mnist-F1-sequential-val
        ] {data/F1.data};
        \addlegendentry{sequential V}

        \addplot[
            dashed, color=MidnightBlue, mark=o,
        ] table [
            x=x-mnist-F1-parallel-val, y=y-mnist-F1-parallel-val
        ] {data/F1.data};
        \addlegendentry{parallel V}
    \end{semilogxaxis}
\end{tikzpicture}
        \caption{\mnist{}}\label{fig:loss-vs-convs-mnist}
    \end{subfigure}
    \hfill
    \begin{subfigure}{\columnwidth}
        \tikzsetnextfilename{F1-cifar10}
\begin{tikzpicture}
    \begin{semilogxaxis}[
        default,
        xlabel={Number of convolutional layers},
        ylabel={Loss},
        xtick={1,2,4,8,16,32},
        xticklabels={1,2,4,8,16,32},
        ]
        \addplot[
            solid, color=BrickRed, mark=x,
        ] table [
            x=x-cifar10-F1-sequential-train, y=y-cifar10-F1-sequential-train
        ] {data/F1.data};
        \addlegendentry{sequential T}

        \addplot[
            solid, color=MidnightBlue, mark=x,
        ] table [
            x=x-cifar10-F1-parallel-train, y=y-cifar10-F1-parallel-train
        ] {data/F1.data};
        \addlegendentry{parallel T}

        \addplot[
            dashed, color=BrickRed, mark=o,
        ] table [
            x=x-cifar10-F1-sequential-val, y=y-cifar10-F1-sequential-val
        ] {data/F1.data};
        \addlegendentry{sequential V}

        \addplot[
            dashed, color=MidnightBlue, mark=o,
        ] table [
            x=x-cifar10-F1-parallel-val, y=y-cifar10-F1-parallel-val
        ] {data/F1.data};
        \addlegendentry{parallel V}
    \end{semilogxaxis}
\end{tikzpicture}
        \caption{\cifar{}}\label{fig:loss-vs-convs-cifar10}
    \end{subfigure}
    \\[2ex]
    \caption{Sequential vs.\ parallel architecture: loss dependence on the number of residual convolutional layers (with a single filter per layer) for the two datasets \mnist{} (left) and \cifar{} (right)}\label{fig:loss-vs-convs}
\end{figure*}

\subsection{With a single filter}\label{sec:experiments:single}
The losses after the \num{100} epochs for the training set (\emph{T}) and the validation set (\emph{V}) are given in~\cref{fig:loss-vs-convs}.
The performance of both architectures can be observed by the points on the red (sequential architecture) and blue (parallel variant) points.
The solid lines represent the training loss and the dashed lines the validation loss.

Due to their identical layout and equal random initialization, training the two networks with one convolutional layer and one filter each resulted consequently in equal loss values.

It can be observed that both architectures perform similarly, in particular for the largest depths of $ 16 $ and $ 32 $.
For \mnist{}, the shallow, parallel architecture slightly outperforms the original, sequential one, while the relationship is inverse for the \cifar{} dataset.

\subsection{With multiple filters}\label{sec:experiments:multiple}
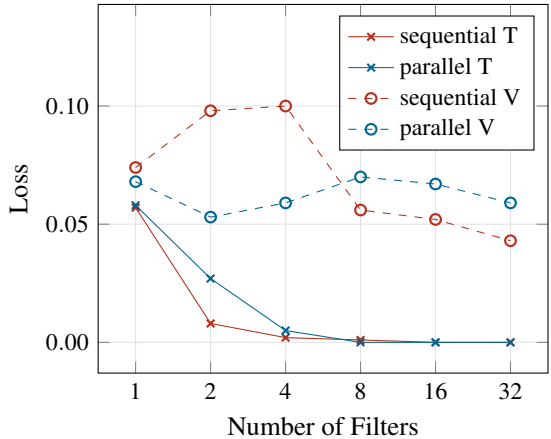
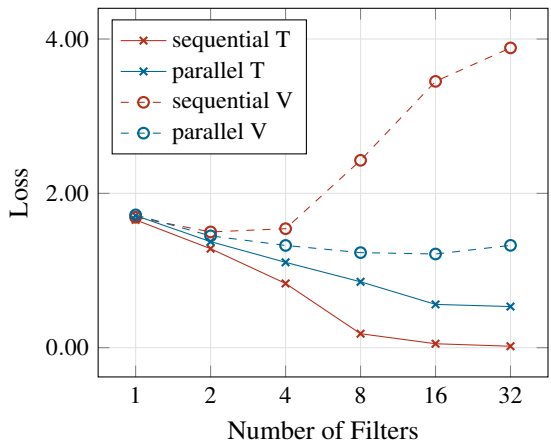
\begin{figure}
    \centering
    \begin{subfigure}{\columnwidth}
        \tikzsetnextfilename{FN-mnist}
\begin{tikzpicture}
    \begin{semilogxaxis}[
        default,
        xlabel={Number of Filters},
        ylabel={Loss},
        xtick={1,2,4,8,16,32},
        xticklabels={1,2,4,8,16,32},
        ymax=0.13,
        legend columns=1,
        legend pos=north east,
    ]
        \addplot[
            solid, color=BrickRed, mark=x,
        ] table [
            x=x-mnist-FN-sequential-train, y=y-mnist-FN-sequential-train
        ] {data/FN.data};
        \addlegendentry{sequential T}

        \addplot[
            solid, color=MidnightBlue, mark=x,
        ] table [
            x=x-mnist-FN-parallel-train, y=y-mnist-FN-parallel-train
        ] {data/FN.data};
        \addlegendentry{parallel T}

        \addplot[
            dashed, color=BrickRed, mark=o,
        ] table [
            x=x-mnist-FN-sequential-val, y=y-mnist-FN-sequential-val
        ] {data/FN.data};
        \addlegendentry{sequential V}

        \addplot[
            dashed, color=MidnightBlue, mark=o,
        ] table [
            x=x-mnist-FN-parallel-val, y=y-mnist-FN-parallel-val
        ] {data/FN.data};
        \addlegendentry{parallel V}
    \end{semilogxaxis}
\end{tikzpicture}
        \caption{\mnist{}}\label{fig:loss-vs-filters-mnist}
    \end{subfigure}
    \\[2ex]
    \begin{subfigure}{\columnwidth}
        \tikzsetnextfilename{FN-cifar10}
\begin{tikzpicture}
    \begin{semilogxaxis}[
        default,
        xlabel={Number of Filters},
        ylabel={Loss},
        xtick={1,2,4,8,16,32},
        xticklabels={1,2,4,8,16,32},
        ymax=4,
        legend columns=1,
        legend pos=north west,
    ]
        \addplot[
            solid, color=BrickRed, mark=x,
        ] table [
            x=x-cifar10-FN-sequential-train, y=y-cifar10-FN-sequential-train
        ] {data/FN.data};
        \addlegendentry{sequential T}

        \addplot[
            solid, color=MidnightBlue, mark=x,
        ] table [
            x=x-cifar10-FN-parallel-train, y=y-cifar10-FN-parallel-train
        ] {data/FN.data};
        \addlegendentry{parallel T}

        \addplot[
            dashed, color=BrickRed, mark=o,
        ] table [
            x=x-cifar10-FN-sequential-val, y=y-cifar10-FN-sequential-val
        ] {data/FN.data};
        \addlegendentry{sequential V}

        \addplot[
            dashed, color=MidnightBlue, mark=o,
        ] table [
            x=x-cifar10-FN-parallel-val, y=y-cifar10-FN-parallel-val
        ] {data/FN.data};
        \addlegendentry{parallel V}
    \end{semilogxaxis}
\end{tikzpicture}
        \caption{\cifar{}}\label{fig:loss-vs-filters-cifar10}
    \end{subfigure}
    \\[2ex]
    \caption{Sequential vs.\ parallel architecture: loss dependence on the number of filters (with 16 convolutional layers) for the two datasets \mnist{} (left) and \cifar{} (right)}\label{fig:loss-vs-filters}
\end{figure}

A single-filter architecture is the most transparent one but it is scarcely used.
It is mostly assumed that more filters are necessary to reach the desired classification performance.
Therefore, experiments with multiple (\numrange{1}{32}) filters per convolutional layer are included.

Same as before, the results after training for \num{100} epochs are shown in~\cref{fig:loss-vs-filters-mnist,fig:loss-vs-filters-cifar10}.
They show an interesting development for \cifar{}: the training loss decreases by raising the number of filters while the validation loss largely increases for more than four filters.
The validation loss considerably deteriorates for the sequential architecture.
(The results for \mnist{} are similar for the training set but less interpretable for the validation set.)

The reason for the distinct picture on \cifar{} is to be sought in relationships between constraints imposed by the task and the number of free trainable parameters~\cite[Chapter~4]{hrycej2023MathematicalFoundationsData}.
A task with $ K = \num{50 000} $ training examples constitutes equally many constraints (resulting from the goal to accurately match the target values) for each output value.
For 10 classes, there are $ M = 10 $ such output values whose reference values are to be correctly predicted by the classifier.
This creates $ K M $ constraints (here: $ \num{50 000} \times 10 = \num{500 000}$).
For the mapping represented by the network, there are $ P $ free (i.e., mutually independent) parameters to make the mapping satisfy the constraints.

\begin{itemize}
    \item
        With $ P = KM $, the system is perfectly determined and could be solved exactly.
    \item
        With $ P > KM $, the system is underdetermined.
        A part of the parameters is set to arbitrary values so that novel examples from the validation set receive arbitrary predictions.
    \item
        With $ P < KM $, the system is overdetermined, and not all constraints can be satisfied.
        This may be useful if the data are noisy, as it is not desirable to fit to noise.
\end{itemize}

An appropriate characteristic is the overdetermination ratio $ Q $ from~\cite{hrycej2022NumberAttentionHeads} defined as
\begin{equation} \label{eq:overdet_ratio}
    Q = \frac{KM}{P}
\end{equation}
The number of genuinely free parameters is difficult to figure out.
It can only be approximated by the total number of parameters, keeping in mind that the number of actually free parameters can be lower.

In training a model by fitting to data, the presence of the noise has to be considered.
The model should reflect the underlying genuine laws in the data but not the noise.
Fitting to the latter is undesirable and is the substance of the well-known phenomenon of \emph{overfitting}.
It was shown in~\cite[Chapter~4]{hrycej2023MathematicalFoundationsData} that fitting to the additive noise and thus the influence of training set noise to the model prediction is reduced to the fraction $ \nicefrac{1}{Q} $.
In other words, it is useful to keep the overdetermination ratio $ Q $ significantly over \num{1}.

\begin{table}
    \centering
    \small
    \caption{Overdetermination ratios for both datasets and different model sizes based on the number of filters per convolutional layer}\label{tab:overdetermination-ratios}
    \begin{tabular}{r rrr}
        \toprule
                  & \multicolumn{3}{r}{Overdetermination ratio $ Q $} \\
        \#filters & \#parameters &     \mnist{} &            \cifar{} \\
        \midrule
              $1$ & \num{14}k    & \num{41.771} &        \num{34.804} \\
              $2$ & \num{37}k    & \num{16.256} &        \num{13.545} \\
              $4$ & \num{106}k   &  \num{5.630} &         \num{4.691} \\
              $8$ & \num{344}k   &  \num{1.743} &         \num{1.453} \\
             $16$ & \num{1.2}M   &  \num{0.495} &         \num{0.412} \\
             $32$ & \num{4.5}M   &  \num{0.132} &         \num{0.110} \\
        \bottomrule
    \end{tabular}
\end{table}

This supplementary information for the plotted variants is given in~\cref{tab:overdetermination-ratios}.
Acceptable values of the overdetermination ratio $ Q $ are given with filter counts of 1, 2, and 4.
This is consistent with the finding that overfitting did not take place in single-filter architectures presented in~\cref{sec:experiments:single}.

For 8 filters or more, $ Q $ is close to 1 or even below it.
In this group, the validation loss can grow arbitrarily although the training loss is reduced.
This is the result of arbitrarily assigned values of underdetermined parameters.

Altogether, the parallel architecture shows better performance on the validation set despite the slightly inferior loss on the training set.
This can be attributed rather to the random effects of underdetermined parameters than to the superiority of one or other architecture.
In this sense, both architectures can be viewed as approximately equivalent concerning their representational capacity.

\subsection{Trade-off of the number of filters and the number of layers}\label{sec:experiments:tradeoff}
As an additional view to the relationship between the depth and the width of the network, a group of experiments is analyzed in which the product of the number of filters ($F$) and the number of convolution layers ($C$) are kept constant.
In this way, also \enquote{intermediary} architectures between deep and shallow ones are captured.
For example, an architecture with $ 32 $ filters and a single convolutional layer has a ratio of $ \nicefrac{1}{32} $ while the ratio with one filter and $ 32 $ layers is $ \nicefrac{32}{1} $.
For $ 16 $ layers with each $ 8 $ filters, it is $ \nicefrac{16}{8} = 2 $.

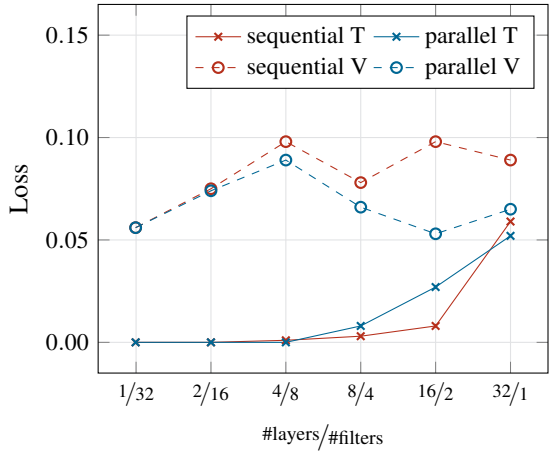
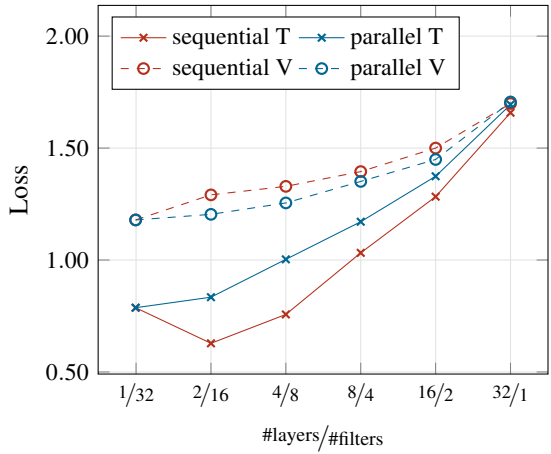
\begin{figure}
    \centering
    \begin{subfigure}{\columnwidth}
        \tikzsetnextfilename{CxF-mnist}
\begin{tikzpicture}
    \begin{semilogxaxis}[
        default,
        xlabel={$\nicefrac{\#\text{layers}}{\#\text{filters}}$},
        ylabel={Loss},
        ymax=0.15,
        legend columns=2,
        legend pos=north east,
        xtick={0.03125, 0.125, 0.5, 2, 8, 32},
        xticklabels={$ \nicefrac{1}{32} $, $ \nicefrac{2}{16} $, $ \nicefrac{4}{8} $, $ \nicefrac{8}{4} $, $ \nicefrac{16}{2} $, $ \nicefrac{32}{1} $},
    ]
        \addplot[
            solid, color=BrickRed, mark=x,
        ] table [
            x=x-mnist-CxF-sequential-train, y=y-mnist-CxF-sequential-train
        ] {data/CxF.data};
        \addlegendentry{sequential T}

        \addplot[
            solid, color=MidnightBlue, mark=x,
        ] table [
            x=x-mnist-CxF-parallel-train, y=y-mnist-CxF-parallel-train
        ] {data/CxF.data};
        \addlegendentry{parallel T}

        \addplot[
            dashed, color=BrickRed, mark=o,
        ] table [
            x=x-mnist-CxF-sequential-val, y=y-mnist-CxF-sequential-val
        ] {data/CxF.data};
        \addlegendentry{sequential V}

        \addplot[
            dashed, color=MidnightBlue, mark=o,
        ] table [
            x=x-mnist-CxF-parallel-val, y=y-mnist-CxF-parallel-val
        ] {data/CxF.data};
        \addlegendentry{parallel V}
    \end{semilogxaxis}
\end{tikzpicture}
        \caption{\mnist{}}\label{fig:loss-vs-layers-mnist}
    \end{subfigure}
    \\[2ex]
    \begin{subfigure}{\columnwidth}
        \tikzsetnextfilename{CxF-cifar10}
\begin{tikzpicture}
    \begin{semilogxaxis}[
        default,
        xlabel={$\nicefrac{\#\text{layers}}{\#\text{filters}}$},
        ylabel={Loss},
        legend columns=2,
        legend pos=north west,
        ymax=2.0,
        xtick={0.03125, 0.125, 0.5, 2, 8, 32},
        xticklabels={$ \nicefrac{1}{32} $, $ \nicefrac{2}{16} $, $ \nicefrac{4}{8} $, $ \nicefrac{8}{4} $, $ \nicefrac{16}{2} $, $ \nicefrac{32}{1} $},
  ]
        \addplot[
            solid, color=BrickRed, mark=x,
        ] table [
            x=x-cifar10-CxF-sequential-train, y=y-cifar10-CxF-sequential-train
        ] {data/CxF.data};
        \addlegendentry{sequential T}

        \addplot[
            solid, color=MidnightBlue, mark=x,
        ] table [
            x=x-cifar10-CxF-parallel-train, y=y-cifar10-CxF-parallel-train
        ] {data/CxF.data};
        \addlegendentry{parallel T}

        \addplot[
            dashed, color=BrickRed, mark=o,
        ] table [
            x=x-cifar10-CxF-sequential-val, y=y-cifar10-CxF-sequential-val
        ] {data/CxF.data};
        \addlegendentry{sequential V}

        \addplot[
            dashed, color=MidnightBlue, mark=o,
        ] table [
            x=x-cifar10-CxF-parallel-val, y=y-cifar10-CxF-parallel-val
        ] {data/CxF.data};
        \addlegendentry{parallel V}
    \end{semilogxaxis}
\end{tikzpicture}
        \caption{\cifar{}}\label{fig:loss-vs-layers-cifar10}
    \end{subfigure}
    \\[2ex]
    \caption{Sequential vs.\ parallel architecture: loss dependence on the ratio of the numbers of layers and filters (product of the number of layers and the number of filters is fixed at 32) for the two datasets \mnist{} (left) and \cifar{} (right)}\label{fig:loss-vs-layers}
\end{figure}

For the product of $32$, there are the following combinations of $C \times F$:
$1 \times 32$, $2 \times 16$, $4 \times 8$, $8 \times 4$, $16 \times 2$ and $32 \times 1$.
In~\cref{fig:loss-vs-layers}, they are ordered along their depth-width ratio $ \nicefrac{C}{F} $:
$ \nicefrac{1}{32} $, $ \nicefrac{2}{16} $, $ \nicefrac{4}{8} $, $ \nicefrac{8}{4} $, $ \nicefrac{16}{2} $, and $ \nicefrac{32}{1} $.
These architectures are represented by the red curves.

As a reference, the blue curve shows their shallow counterparts.
Those are all single-layer architectures.
They differ only in the number of parameters, consistent with their sequential counterparts represented by the red curve.
The difference in the number of parameters is due to the different sizes of the classification layer following the residual connection sequence.
This classification layer is broader for more filters as its input is larger the more filters there are.

Both the training and validation losses increase with the depth-width ratio, indicating the superiority of the shallow architectures.
However, it is important to note that this comparison may not be completely fair due to the inherent difference in parameter numbers.
Specifically, variants with higher depth-width ratios have a diminishing number of parameters resulting from their smaller number of filters.

In~\cref{fig:loss-vs-layers-mnist,fig:loss-vs-layers-cifar10}, it can be observed that the training loss for flattened alternatives is slightly larger compared to the other architectures.
However, the validation loss for flattened alternatives is smaller, albeit to a moderate extent.

In summary, the deep variants can certainly not be viewed as superior in overall terms.
Both architectures are roughly equivalent, as long as the number of parameters is equal.

\section{\uppercase{Statistics of experiments}}\label{sec:statistics}
In addition to experiment runs selected for the presentation in the previous sections, statistics over all \num{6 912} runs, partitioned into some categories, may be useful to complete the performance picture.
Of course, averaging hundreds to thousands of experiments does not guarantee to reflect all theoretical expectations succinctly; it can only confirm rough trends.

\begin{table*}
\centering
\caption{Mean training and validation loss for sequential and parallel architectures and various determination ratios $Q$ intervals}\label{tab:losses-vs-determination-ratio}
\begin{tabular}{l rr | rr | rr | rr} % chktex 44
    \toprule
               & \multicolumn{2}{c}{$ Q \in \rinterval{0}{1} $} & \multicolumn{2}{c}{$ Q \in \rinterval{1}{3} $} & \multicolumn{2}{c}{$ Q \in \rinterval{3}{10} $} & \multicolumn{2}{c}{$ Q \in \rinterval{10}{\infty} $} \\
               & train   &                                  val & train   &                                  val & train   &                                   val & train   &                                        val \\
    \midrule
    \mnist{}   &                                                                                          \multicolumn{8}{c}{~}                                                                                           \\
    sequential & 0.00013 &                              0.05201 & 0.01702 &                              0.12449 & 0.03620 &                               0.11743 & 0.11246 &                                    0.13550 \\
    parallel   & 0.00009 &                              0.07551 & 0.02679 &                              0.11468 & 0.05238 &                               0.11467 & 0.13310 &                                    0.14900 \\
    \midrule
    \cifar{}   &                                                                                          \multicolumn{8}{c}{~}                                                                                           \\
    sequential & 0.25326 &                              2.03107 & 0.72510 &                              1.31691 & 1.07333 &                               1.34721 & 1.58608 &                                    1.65354 \\
    parallel   & 0.52658 &                              1.32386 & 0.88701 &                              1.24884 & 1.17085 &                               1.34227 & 1.63449 &                                    1.68879 \\
    \bottomrule
\end{tabular}
\end{table*}

This statistical summary is presented in~\cref{tab:losses-vs-determination-ratio}.
The losses for training and validation as well as for sequential and parallel architectures are partitioned into intervals of overdetermination ratio to show the different behavior.

According to the theory, with a growing overdetermination ratio, the discrepancy between training and validation loss becomes smaller.
On the other hand, larger overdetermination ratios imply smaller numbers of free network parameters.
Sometimes, this leads to increased losses from the diminished representation capacity of the network.
For ratios smaller than 1, the validation loss may arbitrarily grow because of underdetermined parameters fitted to training data noise (\emph{overfitting}).
This arbitrary growth may be more or less articulated, depending mostly on random factors.
However, there is always a considerable risk of such poor generalization.

As observed in the individual experiments presented, small discrepancies between training and validation loss are reached for overdetermination ratios larger than $ 3 $ for \cifar{} and larger than $ 10 $ for \mnist{}.
These small discrepancies testify to good generalization capability, expected for large overdetermination ratios.

With $ Q < 1 $, the validation loss deteriorates for \cifar{} data if compared with the $ Q $ of the higher interval.
This is the effect of arbitrary parameter values caused by underdetermination.

To summarize, there is a slight advance of shallow architectures for the validation set (five out of eight categories), and deep architectures are better on the training set.
The training and validation losses are mostly closer together for the parallel architecture.

\section{\uppercase{Conclusion}}\label{sec:conclusion}
It is stated in~\cref{sec:decomposition} that a deep residual connection network can be approximately expanded into a sum of shorter (i.e., less deep) sequences of different orders.
Truncating the expansion to the first two terms results in a shallow architecture with a single layer.
This suggests a hypothesis that the representational capacity of such a shallow architecture may be roughly as large as that of the original deep architecture.
If validated, this hypothesis could open avenues to bypass issues typically associated with deep architectures.

Subsequent computational experiments conducted on two widely recognized image classification tasks, \mnist{} and \cifar{}, seem to confirm this theoretically founded expectation.
The performance of both architectures (in configurations with identical numbers of network parameters) is close to each other, with a slight advance of shallow architectures in terms of loss on the validation set.

While the deep architecture performed marginally better on the training set, the cause of its underperformance on the validation set remains an open question.
It is plausible that the deep architecture's ability to capture abrupt nonlinearities may also make it prone to overfitting to noise.
In contrast, the shallow network, due to its inherent smoothness, might exhibit a higher tolerance towards training set noise.

In conclusion, our results suggest a potential parity in the performance of deep and shallow architectures.
It is important to note that the optimization algorithm utilized in this study is a first-order one, which lacks guaranteed convergence properties.
Future research could explore the application of more robust second-order algorithms, which, while not commonly implemented in prevalent software packages, could yield more pronounced results.
This work serves as a preliminary step towards reevaluating architectural decisions in the field of neural networks, urging further exploration into the comparative efficacy of shallow and deep architectures.

\bibliographystyle{apalike}
{\small \bibliography{references}}

\end{document}